\documentclass{article}

\usepackage[final]{neurips_2019}

\usepackage{amsmath,amsfonts,bm}

\def\eqref#1{equation~\ref{#1}}

\def\1{\bm{1}}

\DeclareMathAlphabet{\mathsfit}{\encodingdefault}{\sfdefault}{m}{sl}
\SetMathAlphabet{\mathsfit}{bold}{\encodingdefault}{\sfdefault}{bx}{n}

\DeclareMathOperator{\sign}{sign}

\usepackage[utf8]{inputenc} 
\usepackage[T1]{fontenc}    
\usepackage{hyperref}       
\usepackage{url}            
\usepackage{booktabs}       
\usepackage{amsfonts}       
\usepackage{nicefrac}       
\usepackage{microtype}      
\usepackage{setspace}
\usepackage{algorithm}
\usepackage{algorithmic}
\usepackage{afterpage}
\usepackage{pgfplots}
\pgfplotsset{compat=1.12}
\pgfplotsset{
    dirac/.style={
        mark=triangle*,
        mark options={scale=2},
        ycomb,
        scatter,
        visualization depends on={y/abs(y)-1 \as \sign},
        scatter/@pre marker code/.code={\scope[rotate=90*\sign,yshift=-2pt]}
    }
}

\def \real{\rm I\!R}

\pgfplotsset{
    axis line style={gray}
  }
\title{Regularized Binary Network Training}

\author{%
  Sajad Darabi \\
  Department of Computer Science\\
  University of California Los Angeles\\
  \texttt{sajad.darabi@cs.ucla.edu} \\
   \And
   Mouloud Belbahri  \\
  Department of Mathematics and Statistics\\
  Universit\'e de Montr\'eal  \\
  \texttt{belbahrim@dms.umontreal.ca}
   \AND
   Matthieu Courbariaux \\
   Montreal Institute for Learning Algorithms (Mila) \\
   Universit\'e de Montr\'eal  \\
   \texttt{matthieu.courbariaux@gmail.com}
   \And
   Vahid Partovi Nia\\
   Huawei Noah's Ark Lab \\
   Montreal Research Centre \\
   \texttt{vahid.partovinia@huawei.com}
}

\begin{document}

\maketitle

\begin{abstract}
There is a significant performance gap between Binary Neural Networks (BNNs) and floating point Deep Neural Networks  (DNNs). We propose to improve the binary training method, by introducing a new regularization function that encourages training weights around binary values. In addition, we add trainable scaling factors to our regularization functions. Additionally, an improved approximation of the derivative of the $\sign$ activation function in the backward computation. These modifications are based on linear operations that are easily implementable into the binary training framework. Experimental results on ImageNet shows our method outperforms the traditional BNN method and XNOR-net.
\end{abstract}

\section{Introduction}

DNNs are heavy to compute on low resource devices. There have been several approaches developed to overcome this issue, such as network pruning \citep{lecun1990optimal}, architecture design \citep{Sandler2018InvertedRA}, and quantization \citep{courbariaux2015binaryconnect, Han2015DeepCC}. In particular, weight compression using quantization can achieve very large savings in memory, where binary (1-bit), and ternary (2-bit) approaches have been shown to obtain competitive accuracy as compared to their full precision counterpart \citep{hubara2016binarized, Zhu2016TrainedTQ, tang2017train}. 


Our contribution consists of three ideas that can be easily implemented in the binary training framework presented by \citet{hubara2016binarized} to improve convergence and generalization of binary neural networks. 
First, we improve the straight-through estimator introduced in \cite{hubara2016binarized}, 
Second, we propose a new regularization function that encourage training weights around binary values. Third, a scaling factor is introduced in the regularization function as well as network building blocks to mitigate accuracy drop due to hard binarization. 

Training a binary neural network faces two major challenges: quantizing the weights, and the activation functions. As both weights and activations are binary, the traditional continuous optimization methods such as SGD cannot be directly applied. Instead, a continuous approximation is used for the $\sign$ activation during the backward pass. 

\begin{figure}[htb]
    \centering
    \includegraphics[scale=0.7]{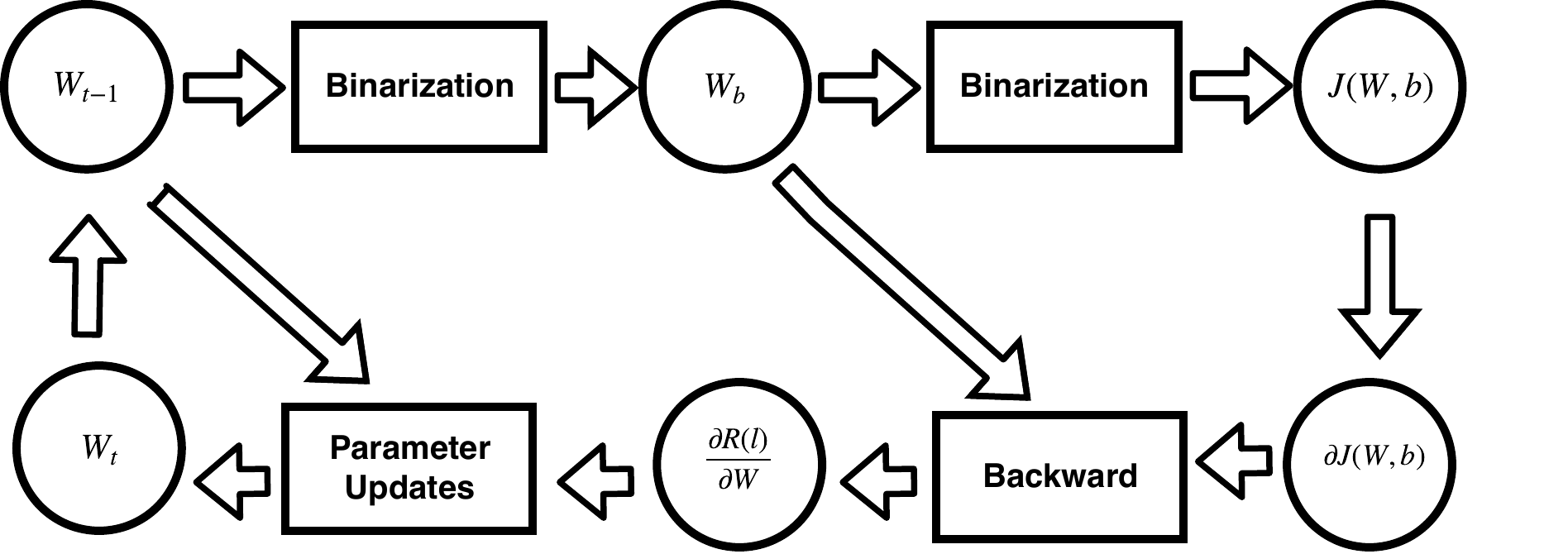}
    \caption{Binary training where arrows indicate operands flowing into operations or blocks.}
  \label{fig:bintraining}
\end{figure}

The general training framework is depicted in Figure~\ref{fig:bintraining}. In \cite{hubara2016binarized}, the weights are quantized by using the $\sign$ function which is $+1$ if $w > 0$ and $-1$ otherwise.

Binary training use heuristics  to  approximating the gradient of a neuron, 
\begin{align}
    \frac{d L(w)}{d w} \approx \frac{d L}{d w} \bigg\rvert_{w=w^b} \1_{\{|w| \leq 1\}}
    \label{lossderivative}
\end{align}
where $L$ is the loss function, $\1(.)$ is the indicator function and $w^b$ is the binarized weight. The gradients in the backward pass are then applied to weights that are within $[-1, +1]$. The training process is summarized in Figure \ref{fig:bintraining}. As weights undergo gradient updates, they are eventually pushed out of the center region and instead make two modes, one at $-1$ and another at $+1$.

\section{Gradient Approximation}

\begin{figure}[htb]
    \begin{center}
    \begin{tikzpicture}[scale=0.45]
        \begin{axis}[
            axis lines=middle,
            ylabel={$\sign(x)$},
            ylabel style={font=\Huge, yshift=3pt},
            xmin=-3, xmax=3,
            ymin=-2, ymax=2,
            xtick=\empty,
            ytick={0, 1},
            extra y ticks={-1},
            extra y tick style={
              tick label style={anchor=west, xshift=3pt},
            },
            function line/.style={
              black,
              thick,
              samples=2,
            },
            single dot/.style={
              red,
              mark=*,
            },
            empty point/.style={
              only marks,
              mark=*,
              mark options={fill=white, draw=black},
            },
          ]
            \addplot[function line, domain=\pgfkeysvalueof{/pgfplots/xmin}:0] {-1};
            \addplot[function line, domain=0:\pgfkeysvalueof{/pgfplots/xmax}] {1};
            \addplot[empty point] coordinates {(0, -1) (0, 1)};
        \end{axis}
    \end{tikzpicture} 
    \hspace{0.1cm}
    \begin{tikzpicture}[scale=0.45]
        \begin{axis}[
            axis lines=middle,
            ylabel={$\delta(x)$},
            ylabel style={font=\Huge, yshift=3pt},
            xmin=-3, xmax=3,
            ymin=-2, ymax=2,
            xtick=\empty,
            ytick=\empty,
            extra y ticks={1},
            extra y tick style={
              tick label style={anchor=west, yshift=5pt},
            },
            function line/.style={
              black,
              thick,
              samples=2,
            },
            single dot/.style={
              red,
              mark=*,
            },
            empty point/.style={
              only marks,
              mark=*,
              mark options={fill=white, draw=black},
            },
          ]
          \draw (0,0) -- (0,1);
          \draw (0,0) -- (2,0);
          \draw (0,0) -- (-2,0);
        \end{axis}
    \end{tikzpicture}
        \hspace{.1cm}
    \begin{tikzpicture}[scale=0.45]
        \begin{axis}[
            axis lines=middle,
            ylabel={$\mathrm{htanh}(x)$},
            ylabel style={font=\Huge, yshift=3pt},
            xmin=-3, xmax=3,
            ymin=-2, ymax=2,
            xtick=\empty,
            ytick={0, 1},
            extra y ticks={-1},
            extra y tick style={
              tick label style={anchor=west, xshift=3pt},
            },
            function line/.style={
              black,
              thick,
              samples=2,
            },
            single dot/.style={
              red,
              mark=*,
            },
            empty point/.style={
              only marks,
              mark=*,
              mark options={fill=white, draw=black},
            },
          ]
            \addplot[function line, domain=\pgfkeysvalueof{/pgfplots/xmin}:-1] {-1};
            \addplot[function line, domain=1:\pgfkeysvalueof{/pgfplots/xmax}] {1};
            \addplot[black, domain=-1:1, smooth]{x};
        \end{axis}
    \end{tikzpicture} 
    \hspace{0.1cm}
    \begin{tikzpicture}[scale=0.45]
        \begin{axis}[
            axis lines=middle,
            ylabel={$\mathrm{STE}(x)$},
            ylabel style={font=\Huge, yshift=3pt},
            xmin=-3, xmax=3,
            ymin=-2, ymax=2,
            xtick=\empty,
            ytick=\empty,
            extra y ticks={1},
            extra y tick style={
              tick label style={anchor=west, yshift=5pt},
            },
            function line/.style={
              black,
              thick,
              samples=2,
            },
            single dot/.style={
              red,
              mark=*,
            },
            empty point/.style={
              only marks,
              mark=*,
              mark options={fill=white, draw=black},
            },
          ]
            \addplot[function line, domain=-1:1] {1};
            \addplot [mark=none] coordinates {(1, 1) (1, 0)};
            \addplot [mark=none] coordinates {(-1, 0) (-1, 1)};
        \end{axis}
    \end{tikzpicture}
        \vspace{.1cm}
    \begin{tikzpicture}[scale=0.45]
            \begin{axis}[
      axis x line=middle, axis y line=middle, samples = 100,  ylabel={$\mathrm{SS}_{1}(x)$},ylabel style={font=\Huge, yshift=3pt}, ymin=-2, ymax=2, ytick={-1,0,1}, xtick=\empty,
      xmin=-4, xmax=4
    ]
    \addplot[black, domain=-4:4, smooth]{2*(1+pow(e,-1*x))^(-1)*(1+1*x*pow(e,-1*x)*(1+pow(e,-1*x))^(-1)))-1};
    \end{axis}
    \end{tikzpicture} 
    \begin{tikzpicture}[scale=0.45]
    \begin{axis}[
      axis x line=middle, axis y line=middle, samples = 100,  ylabel={$\frac{d\mathrm{SS}_{1}(x)}{dx}$}, ylabel style={font=\Huge, yshift=3pt},ymin=-2, ymax=2, ytick={-1,0,1}, xtick=\empty,
      xmin=-4, xmax=4
    ]
    \addplot[black, domain=-4:4, smooth]{2 * (-1^2 * x * pow(e, -1*x) * (1+pow(e,-1*x))^(-2) + 2 * 1^2 * x * pow(e, -2*1*x) * (1+pow(e,-1*x))^(-3) + 2 * 1 * pow(e, -1*x) * (1+pow(e,-1*x))^(-2)) };
    \end{axis}
    \end{tikzpicture}
    \hspace{.1cm}
        \begin{tikzpicture}[scale=0.45]
            \begin{axis}[
      axis x line=middle, axis y line=middle, samples = 100,  ylabel={$\mathrm{SS}_{5}(x)$},ylabel style={font=\Huge, yshift=3pt}, ymin=-2, ymax=2, ytick={-1,0,1}, xtick=\empty,
      xmin=-4, xmax=4
    ]
    \addplot[black, domain=-4:4, smooth]{2*(1+pow(e,-5*x))^(-1)*(1+5*x*pow(e,-5*x)*(1+pow(e,-5*x))^(-1)))-1};
    \end{axis}
    \end{tikzpicture} 
    \hspace{0.1cm}
    \begin{tikzpicture}[scale=0.45]
    \begin{axis}[
      axis x line=middle, axis y line=middle, samples = 100,  ylabel={$\frac{d\mathrm{SS}_{5}(x)}{dx}$}, ylabel style={font=\Huge, yshift=3pt},ymin=-2, ymax=6, ytick=\empty, xtick=\empty,
      xmin=-4, xmax=4
    ]
    \addplot[black, domain=-4:4, smooth]{2 * (-5^2 * x * pow(e, -5*x) * (1+pow(e,-5*x))^(-2) + 2 * 5^2 * x * pow(e, -2*5*x) * (1+pow(e,-5*x))^(-3) + 2 * 5 * pow(e, -5*x) * (1+pow(e,-5*x))^(-2)) };
    \end{axis}
    \end{tikzpicture}
    \end{center}
    \caption{Forward and backward approximations. The forward and backward functions of SignSwish  $SS_\beta(x)$ for $\beta=1$. and $\beta=5$.}
    \label{fig:derivativeapprox}
\end{figure}
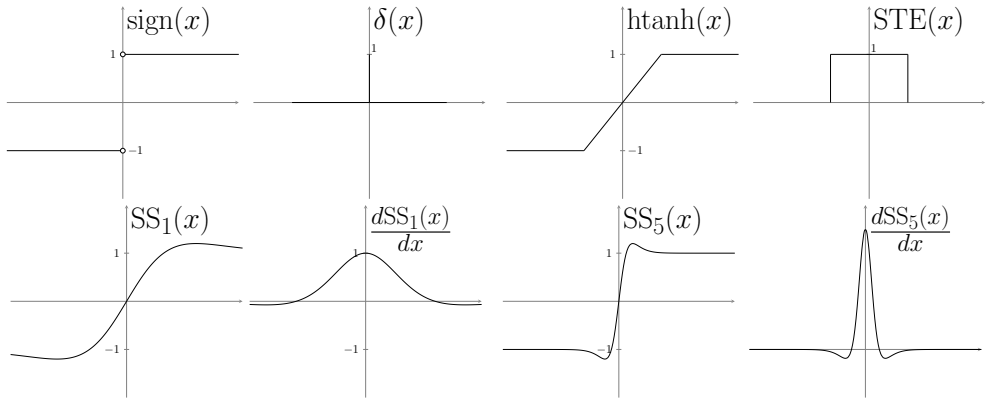

 Our first modification is on closing the discrepancy between the forward pass and backward pass. Originally, the $\sign$ function derivative is approximated using the $\mathrm{htanh}(x)$ activation derivative as shown in Figure \ref{fig:derivativeapprox}. Instead, we modify the Swish-like activation \citep{Ramachandran2017SearchingFA, elfwing2018sigmoid, Hendrycks2016BridgingNA}, which has been shown to outperform other activation functions on various tasks. The modifications are performed by taking its derivative and centering it around 0. Let $\mathrm{SS}_{\beta}(x) = 2 \sigma(\beta x) \left[ 1 + \beta x \{1- \sigma(\beta x)\} \right] - 1,$
where $\sigma(z)$ is the sigmoid function and the scale $\beta > 0$ controls how fast the activation function asymptotes to $-1$ and $+1$. The $\beta$ parameter can be learned by the network or be hand-tuned as a hyperparameter. As opposed to the Swish function, where it is unbounded on the right side, the modification makes it bounded, a more valid approximator of the $\sign$ function so we call this activation SignSwish or SSwish, see Figure~\ref{fig:derivativeapprox}.

\citet{hubara2016binarized} noted that the Straight-through Estimator (STE) fails to learn weights near the borders of $-1$ and $+1$. As depicted in Figure \ref{fig:derivativeapprox}, our proposed SignSwish activation alleviates this issue, as it remains differentiable near $-1$ and $+1$ allowing weights to change signs during training if necessary.

Note that the derivative $\frac{d\mathrm{SS}_{\beta}(x)}{dx}$ is zero at two points, controlled by $\beta$. Indeed, it is simple to show that the derivative is zero for $x \approx \pm 2.4 / \beta$. By adjusting this parameter, it is possible to adjust the location at which the gradients start saturating, in contrast with the STE estimators where it is fixed.

\section{Regularization function}
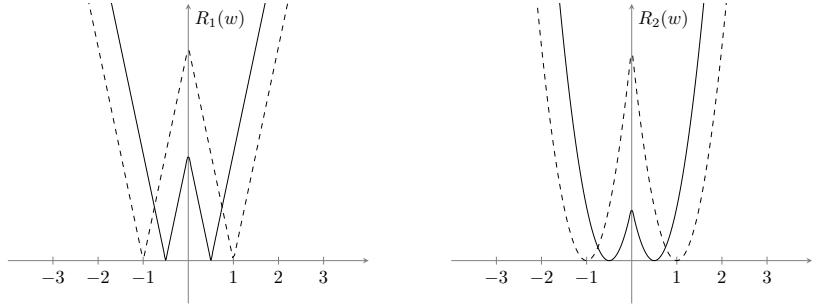
\begin{figure}[htb]
    \begin{center}
    \begin{tikzpicture}[scale=0.7]
    \begin{axis}[
      axis x line=middle, ylabel={$R_1(w)$}, axis y line=middle, samples = 200,
      ymin=-0.2, ymax=1.2, ytick={0},
      xmin=-4, xmax=4, xtick={-3,...,3}
    ]
    \addplot[black, domain=-4:4, dashed]{abs(1-abs(x))};
    \addplot[black, domain=-4:4, smooth]{abs(0.5-abs(x))};
    \end{axis}
    \end{tikzpicture} 
    \hspace{1cm}
    \begin{tikzpicture}[scale=0.7]
    \begin{axis}[
      axis x line=middle, ylabel={$R_2(w)$}, axis y line=middle, samples = 200,
      ymin=-0.2, ymax=1.2, ytick={0},
      xmin=-4, xmax=4, xtick={-3,...,3}
    ]
    \addplot[black, domain=-4:4, dashed]{(1-abs(x))^2};
    \addplot[black, domain=-4:4, smooth]{(0.5-abs(x))^2};
    \end{axis}
    \end{tikzpicture}
    \end{center}
    \caption{Regularization functions for $\alpha = 0.5$ (solid line) and $\alpha = 1$ (dashed line). }
    \label{fig:regfun}
\end{figure}

In general, a regularization term is added to a model to prevent over-fitting and to obtain robust generalization. The two most commonly used regularization terms are $L_1$ and $L_2$ norms. If one were to embed these regularization functions in binary training, it would encourage the weights to be near zero, though this does not align with the objective of a binary network. A regularization function for binary networks should vanish upon the quantized values. Following this intuition, we define a function that encourages the weights around $-1$ and $+1$.  The Manhattan $R_1(.)$ and Euclidean $R_2(.)$ regularization functions are defined as

\begin{align}
    R_{1}(w) = |\alpha - |w||,\quad     R_{2}(w) = (\alpha - |w|)^2,
    \label{l1reg}
\end{align}
where $\alpha \in \real^+$. In Figure \ref{fig:regfun}, we depict the different regularization terms to help with intuition.

These regularization functions encode a quantization structure, when added to the overall loss function of the network encouraging weights to binary values. The difference between the two is in the rate at which weights are penalized when far from the binary objective, $L_1$ linearly penalizes the weights and is non-smooth compared to the $L_2$ version where weights are penalized quadritically. We further relax the hard thresholding of binary values $\{-1, 1\}$ by introducing scales $\alpha$ in the regularization function. This results in a symmetric regularization function with two minimums, one at $-\alpha$ and another at $+\alpha$. The scales are then added to the networks and multiplied into the weights after the binarization operation. As these scales are introduced in the regularization function and are embedded into the layers of the network they can be learned using back-propagation. This is in contrast with the scales introduced in \cite{rastegari2016xnor}, where they compute the scales dynamically during training using the statistics of the weights after every training batch.
As depicted in Figure \ref{fig:regfun}, in the case of $\alpha = 1$, the weights are penalized at varying degrees upon moving away from the objective quantization values, in this case, $\{-1, +1\}$. 


\section{Training Procedure}
Combining both the regularization and modified STE ideas, we adapt the training procedure by replacing the $\sign$ backward approximation with that of the derivative of $\mathrm{SS}_{\beta}$ activation (\ref{sswish}). During training, the real weights are no longer clipped as in BNN training, as the network can back-propagate through the $\mathrm{SS}_{\beta}$ activation and update the weights correspondingly. 

Additional scales are introduced to the network, which multiplies into the weights of the layers. The regularization terms introduced are then added to the total loss function,
\begin{align}
    J(W, b) = L(W, b) + \lambda_t \sum_{h} R(W_h,\alpha_h),
\end{align}
where $L(W, b)$ is the cost function, $W$ and $b$ are the sets of all weights and biases in the network, $W_h$ is the set weights at layer $h$ and $\alpha_h$ is the corresponding scaling factor. 


The scale $\alpha$ is a single scalar per layer, or as proposed in \citet{rastegari2016xnor} is a scalar for each filter in a convolutional layer. For example, given a CNN block with weight dimensionality $(C_{\mathrm{in}}, C_{\mathrm{out}}, H, W)$, where $C_{\mathrm{in}}$ is the number of input channels, $C_{\mathrm{out}}$ is the number of output channels, and $H$, $W$, the height and width of the filter respectively, then the scale parameter would be a vector of dimension $C_{\mathrm{out}}$, that factors into each filter. 


As the scales are learned jointly with the network through back-propagation, it is important to initialize them appropriately. In the case of the Manhattan penalizing term (\ref{l1reg}), given a scale factor $\alpha$ and a weight filter, the objective is to solve
\begin{align}
    \min_\alpha \sum_{h,w} \Bigl\lvert \alpha - |W_{h,w}|\Bigr\rvert, \quad \min_\alpha \sum_{h,w} \left(\alpha - |W_{h,w}|\right)^2
\end{align}
The minimum of the above functions are
\begin{align}
    \alpha^* = \mathrm{median}(|W|), \quad  \alpha^* = \mathrm{mean}(|W|)
\end{align}
depending if the Manhattan or Euclidean regularization is used (\ref{l1reg}).
The Euclidean regularization coincides with the  scaling factor derived in \citet{rastegari2016xnor}. The difference here is that we have the choice to embed it in back-propagation in our framework, as opposed to computing the values dynamically. 

\section{Experimental Results}
\label{experiments}
\begin{table}[htb]
\caption{Top-1 and top-5 accuracies (in percentage) on ImageNet dataset, of different combinations of the proposed technical novelties on different architectures. 
}
\label{tab:ablation}
\begin{center}
\resizebox{8cm}{!}{
\begin{tabular}{lll cccccc}
 &\multicolumn{1}{l}{\bf Reg.} &\multicolumn{1}{l}{\bf Activation} &\multicolumn{2}{c}{\bf AlexNet} &\multicolumn{2}{c}{\bf Resnet-18} 
\\ \hline \\
&&& Top-1 & Top-5 & Top-1 & Top-5 \\\\
       & & $\mathrm{SS}_5$  & 46.11 & 75.70 & 52.64 & 72.98 \\
       &\raisebox{0.5ex}{$R_1$} & $\mathrm{SS}_{10}$ & 46.08 & 75.75 & 51.13 & 74.94\\
       & & $\mathrm{htanh}$ & 41.58 & 69.90 & 50.72 & 73.48 \\[1ex]
       & & $\mathrm{SS}_5$ & 45.62 & 70.13 & 53.01&  72.55 \\
       &\raisebox{0.5ex}{$R_2$}  & $\mathrm{SS}_{10}$ & 45.79 & 75.06 & 49.06 & 70.25  \\
       & & $\mathrm{htanh}$ & 40.68 & 68.88 & 48.13 & 72.72 \\[1ex]
       & & $\mathrm{SS}_5$  & 45.25 & 75.30 & 43.23 &68.51 \\
       &\raisebox{0.5ex}{None} & $\mathrm{SS}_{10}$ & 45.60 & 75.30 & 44.50 &64.54 \\
       & & $\mathrm{htanh}$ & 39.18 & 69.88 &  42.46 &  67.56 \\
\end{tabular}
}
\end{center}
\end{table}

\begin{table}[htb]
\caption{Comparison of top-1 and top-5 accuracies of our method with BinaryNet, XNOR-Net and ABC-Net on ImageNet, summarized from Table \ref{tab:ablation}. The results of BNN, XNOR, \& ABC-Net are reported from the corresponding papers \citep{rastegari2016xnor, hubara2016binarized, tang2017train}. Results for ABC-NET on AlexNet were not available, and so is not reported.}
\label{tab:comparison}
\begin{center}

\begin{tabular}{l cccc}
 \multicolumn{1}{c}{\bf Method} &\multicolumn{2}{c}{\bf AlexNet} &\multicolumn{2}{c}{\bf Resnet-18} 
\\ \hline \\
& Top-1 & Top-5 & Top-1 & Top-5 \\\\
\bf Ours & $46.1\%$ & $75.7\%$ & $53.0\%$ & $72.6\%$ \\
BinaryNet & $41.2\%$ & $65.6\%$ & $42.2\%$ & $67.1\%$ \\
XNOR-Net & $44.2\%$ & $69.2\%$ & $51.2\%$ & $ 73.2\%$ \\
ABC-Net & - & - & $42.7\%$ & $67.6\%$ \\\\
Full-Precision & $56.6\%$ & $80.2\%$ & $69.3\%$ & $89.2\%$\\

\end{tabular}
\end{center}
\end{table}

We evaluate the performance of our training method on two architectures: AlexNet and Resnet-18 \citep{he2016deep} on ImageNet \citep{hubara2016binarized, rastegari2016xnor, tang2017train}. Following previous work, we used batch-normalization before each activation function. Additionally, we keep the first and last layers to be in full precision, as we lose $2-3\%$ accuracy otherwise. This approach is followed by other binary methods that we compare to \citep{hubara2016binarized,rastegari2016xnor, tang2017train}. The results are summarized in Table \ref{tab:ablation}. In all the experiments involving $R_1$ regularization we set the $\lambda$ to $10^{-7}$ and $R_2$ regularization in ranges of  $10^{-5}-10^{-7}$. Also, in every network, the scales are introduced per filter in convolutional layers, and per column in fully connected layers. The weights are initialized using a pre-trained model with $\mathrm{htan}$ activation function as done in \cite{Liu2018BiRealNE}. Then the learning rate for AlexNet is set to $2.33\times10^-3$ and multiplied by $0.1$ at the $12^{th}$ and $18^{th}$ epochs for a total of $25$ epochs trained. For the $18$-layers ResNet, the learning rate is started from $0.01$ and multiplied by $0.1$ at the $10^{th}$, $20^{th}$ and $30^{th}$ epochs. On the ImageNet dataset, we run a small ablation study of our regularized binary network training method with fixed $\beta$ parameters.

\bibliographystyle{plainnat}
\bibliography{neurips_2019}

\begin{thebibliography}{13}
\providecommand{\natexlab}[1]{#1}
\providecommand{\url}[1]{\texttt{#1}}
\expandafter\ifx\csname urlstyle\endcsname\relax
  \providecommand{\doi}[1]{doi: #1}\else
  \providecommand{\doi}{doi: \begingroup \urlstyle{rm}\Url}\fi

\bibitem[Courbariaux et~al.(2015)Courbariaux, Bengio, and
  David]{courbariaux2015binaryconnect}
Matthieu Courbariaux, Yoshua Bengio, and Jean-Pierre David.
\newblock Binaryconnect: Training deep neural networks with binary weights
  during propagations.
\newblock In \emph{Advances in neural information processing systems}, pages
  3123--3131, 2015.

\bibitem[Elfwing et~al.(2018)Elfwing, Uchibe, and Doya]{elfwing2018sigmoid}
Stefan Elfwing, Eiji Uchibe, and Kenji Doya.
\newblock Sigmoid-weighted linear units for neural network function
  approximation in reinforcement learning.
\newblock \emph{Neural Networks}, 2018.

\bibitem[Han et~al.(2015)Han, Mao, and Dally]{Han2015DeepCC}
Song Han, Huizi Mao, and William~J. Dally.
\newblock Deep compression: Compressing deep neural network with pruning,
  trained quantization and huffman coding.
\newblock \emph{CoRR}, abs/1510.00149, 2015.

\bibitem[He et~al.(2016)He, Zhang, Ren, and Sun]{he2016deep}
Kaiming He, Xiangyu Zhang, Shaoqing Ren, and Jian Sun.
\newblock Deep residual learning for image recognition.
\newblock In \emph{Proceedings of the IEEE conference on computer vision and
  pattern recognition}, pages 770--778, 2016.

\bibitem[Hendrycks and Gimpel(2016)]{Hendrycks2016BridgingNA}
Dan Hendrycks and Kevin Gimpel.
\newblock Bridging nonlinearities and stochastic regularizers with gaussian
  error linear units.
\newblock \emph{CoRR}, abs/1606.08415, 2016.

\bibitem[Hubara et~al.(2018)Hubara, Courbariaux, Soudry, El-Yaniv, and
  Bengio]{hubara2016binarized}
Itay Hubara, Matthieu Courbariaux, Daniel Soudry, Ran El-Yaniv, and Yoshua
  Bengio.
\newblock Quantized neural networks: Training neural networks with low
  precision weights and activations.
\newblock \emph{Journal of Machine Learning Research}, 18\penalty0
  (187):\penalty0 1--30, 2018.

\bibitem[LeCun et~al.(1990)LeCun, Denker, and Solla]{lecun1990optimal}
Yann LeCun, John~S Denker, and Sara~A Solla.
\newblock Optimal brain damage.
\newblock In \emph{Advances in neural information processing systems}, pages
  598--605, 1990.

\bibitem[Liu et~al.(2018)Liu, Wu, Luo, Yang, Liu, and Cheng]{Liu2018BiRealNE}
Zechun Liu, Baoyuan Wu, Wenhan Luo, Xin Yang, Wei Liu, and Kwang-Ting Cheng.
\newblock Bi-real net: Enhancing the performance of 1-bit cnns with improved
  representational capability and advanced training algorithm.
\newblock In \emph{ECCV}, 2018.

\bibitem[Ramachandran et~al.(2017)Ramachandran, Zoph, and
  Le]{Ramachandran2017SearchingFA}
Prajit Ramachandran, Barret Zoph, and Quoc~V. Le.
\newblock Searching for activation functions.
\newblock \emph{CoRR}, abs/1710.05941, 2017.

\bibitem[Rastegari et~al.(2016)Rastegari, Ordonez, Redmon, and
  Farhadi]{rastegari2016xnor}
Mohammad Rastegari, Vicente Ordonez, Joseph Redmon, and Ali Farhadi.
\newblock Xnor-net: Imagenet classification using binary convolutional neural
  networks.
\newblock In \emph{European Conference on Computer Vision}, pages 525--542.
  Springer, 2016.

\bibitem[Sandler et~al.(2018)Sandler, Howard, Zhu, Zhmoginov, and
  Chen]{Sandler2018InvertedRA}
Mark Sandler, Andrew~G. Howard, Menglong Zhu, Andrey Zhmoginov, and Liang-Chieh
  Chen.
\newblock Inverted residuals and linear bottlenecks: Mobile networks for
  classification, detection and segmentation.
\newblock \emph{CoRR}, abs/1801.04381, 2018.

\bibitem[Tang et~al.(2017)Tang, Hua, and Wang]{tang2017train}
Wei Tang, Gang Hua, and Liang Wang.
\newblock How to train a compact binary neural network with high accuracy?
\newblock In \emph{AAAI}, pages 2625--2631, 2017.

\bibitem[Zhu et~al.(2016)Zhu, Han, Mao, and Dally]{Zhu2016TrainedTQ}
Chenzhuo Zhu, Song Han, Huizi Mao, and William~J. Dally.
\newblock Trained ternary quantization.
\newblock \emph{CoRR}, abs/1612.01064, 2016.

\end{thebibliography}
\end{document}


\title{Supplemental Material}

\maketitle

\section{CIFAR 10 Results}

\begin{table}[htb]
\caption{Top-1 accuracies (in percentage) on CIFAR-10, using regularization functions (\ref{l1reg}) and (\ref{l2reg}) with AlexNet and DoReFa-Net. STE stands for Straigth-Through Estimator and $SS_t$ for SwishSign with a trainable $\beta$. XNOR scales experiments are done using the original XNOR implementation.}
\label{tab:cifar10comparison}
\begin{center}

\begin{tabular}{lll cccc}
 &\multicolumn{1}{l}{\bf Scale} &\multicolumn{1}{l}{\bf STE} &\multicolumn{1}{c}{\bf AlexNet} &\multicolumn{1}{c}{\bf DoReFa-Net\footnotemark} 
\\ \hline \\
&&& Top-1 & Top-1  \\\\
      & & $\mathrm{SS}_{t}$  & 87.16  & 83.92 \\
      &\raisebox{0.5ex}{$R_1$} & $\mathrm{SS}_{10}$ & 86.80  & 83.21 \\
      & & $\mathrm{SS}_{5}$ & 86.85  & 83.72 \\
      & & $\mathrm{tanh}$ & 87.06  & 82.79 \\
      & & $\mathrm{bireal}$ & 87.04  & 82.64 \\
      & & $\mathrm{htanh}$ & 86.90   & 82.78 \\[1ex]
      & & $\mathrm{SS}_t$ & 87.30  & 83.52 \\
      &\raisebox{0.5ex}{$R_2$}  & $\mathrm{SS}_{10}$ & 86.39  & 82.43 \\
      & & $\mathrm{SS}_{5}$ & 86.80  & 83.58 \\
        & & $\mathrm{tanh}$ & 87.13  & 82.33 \\
        & & $\mathrm{bireal}$ & 87.39  & 82.93 \\
      & & $\mathrm{htanh}$ & 87.20  & 82.33 \\[1ex]
        & & $\mathrm{SS}_t$  & 84.94  & 82.58 \\
      &\raisebox{0.5ex}{XNOR\footnotemark} & $\mathrm{SS}_{10}$ & 83.64  & 83.12 \\
      & & $\mathrm{SS}_{5}$ & 85.74  & 82.54 \\
         & & $\mathrm{tanh}$ & 80.30  & 81.68 \\
          & & $\mathrm{bireal}$ & 83.84  & 82.58 \\
      & & $\mathrm{htanh}$ & 83.87  & 82.08 \\ [1ex]
      & & $\mathrm{SS}_t$  & 86.24  & 83.00 \\
      &\raisebox{0.5ex}{None} & $\mathrm{SS}_{10}$ & 84.70  & 82.98 \\
      & & $\mathrm{SS}_{5}$ & 84.85  & 83.14 \\
         & & $\mathrm{tanh}$ & 86.12  & 82.33 \\
          & & $\mathrm{bireal}$ & 86.24  & 82.56 \\
      & & $\mathrm{htanh}$ & 85.97  & 82.51 \\
\end{tabular}

\end{center}
\end{table}

\footnotetext{We follow the implementation \url{https://github.com/allenai/XNOR-Net} and add the various STEs to the implementation}
\footnotetext{We use the SVHN model \url{https://github.com/ppwwyyxx/tensorpack/blob/master/examples/DoReFa-Net/svhn-digit-dorefa.py}}
The CIFAR-10 data \cite{krizhevsky2009learning} consists of $50,000$ train images and a test set of $10,000$. We apply the common data augmentation strategy for CIFAR-10, where images are padded by $4$ pixels on each side, a random $32\times32$ crop is taken, and a horizontal flip with probability $0.5$, which then the image is normalized with the dataset statistics. 

In this section we run an ablation study in order to better understand our proposed additions to the binary training framework. We train both, AlexNet\footnote{For AlexNet we use the image transformations explained in the ImageNet section} \cite{krizhevsky2012imagenet}, and DoReFA-Net \cite{Zhou2016DoReFaNetTL} using the ADAM \cite{Kingma2014AdamAM} optimizer. For both networks batch normalization layers are added prior to activations.

To train AlexNet we use batch size of $64$ for training. The initial learning rates were set to $0.005$ and decayed by $0.1$ on epoch $30$ and $45$ for a total of $55$ epochs. For DoReFa-Net, the initial learning rates were set to $0.001$ and decayed by $0.1$ on epochs $100$, $150$ and $200$ for a total of $250$ epochs. In all of the experiments, we set the regularization parameter $\lambda$ to $5*10^{-7}$. Lastly, the networks are trained from scratch where weights are initialized using \citet{glorot2010understanding} initialization. In the case of $R_1$ and $R_2$, scales are introduced for each convolution filter, and are initialized using the median or the mean of the of the absolute value of the filter weights. The performance of these scales is also compared to the original XNOR-net implementation. Lastly, for our ablation study on the parameter of the SwishSign, we try three different settings: one with trainable parameter $\beta$, and the two others with $\beta=5$ and $\beta=10$ fixed. The results for the ablation study are summarized in Table \ref{tab:cifar10comparison}.

To study the effect of our SwishSign, we compare it to the different backward approximations most recently proposed: bireal and the soft-hinge \cite{Friesen2017DeepLA}. From the DoReFa-Net experiments in Table \ref{tab:cifar10comparison} the $SS_t$ with trainable parameter $\beta$ performs the highest, achieving $83.92 \%$ top-1 accuracy; a $1\%$ accuracy drop compared to the experiments ran in \cite{Friesen2017DeepLA}, with hard thresholding units and real weights. The results of AlexNet on CIFAR-10 also show the same effect, where SwishSign improves the results. From Table \ref{tab:ablation}, and referring to networks without regularization, we see the benefit of using SwishSign approximation versus the STE outperforming it in almost all cases.

\bibliographystyle{plainnat}
\bibliography{neurips_2019}